\documentclass[acmsmall,screen]{acmart}

\usepackage{amsthm}
\usepackage{algorithmicx}
\usepackage{subfigure}
\usepackage{booktabs}
\usepackage{array}
\usepackage[table]{xcolor}
\usepackage{multirow}
\usepackage{supertabular}
\usepackage[linesnumbered,ruled,vlined]{algorithm2e}
\newtheorem{definition}{Definition}

\AtBeginDocument{%
  }

\definecolor{LightGray}{gray}{0.95}
\definecolor{LightBlue}{RGB}{230,245,250}
\definecolor{LightYellow}{RGB}{255,250,220}

\setcopyright{acmlicensed}
\copyrightyear{2025}
\acmYear{2025}
\acmDOI{XXXXXXX.XXXXXXX}
\begin{document}

\title{LMM-Incentive: Large Multimodal Model-based Incentive Design for User-Generated Content in Web 3.0}

\author{Jinbo Wen}
\affiliation{%
  \institution{Nanjing University of Aeronautics and Astronautics}
  \city{Nanjing}
  \country{China}}
\email{jinbo1608@nuaa.edu.cn}

\author{Jiawen Kang}
\affiliation{%
  \institution{Guangdong University of Technology}
  \city{Guangzhou}
  \country{China}}
\email{kavinkang@gdut.edu.cn}

\author{Linfeng Zhang}
\affiliation{%
  \institution{The Hong Kong Polytechnic University}
  \city{Hong Kong}
  \country{Hong Kong}}
\email{lamfung.zhang@connect.polyu.hk}

\author{Xiaoying Tang}
\affiliation{%
  \institution{The Chinese University of Hong Kong}
  \city{Shenzhen}
  \country{China}}
\email{tangxiaoying@cuhk.edu.cn}

\author{Jianhang Tang}
\affiliation{%
  \institution{Guizhou University}
  \city{Guiyang}
  \country{China}}
\email{jhtang@gzu.edu.cn}

\author{Yang Zhang}
\affiliation{%
  \institution{Nanjing University of Aeronautics and Astronautics}
  \city{Nanjing}
  \country{China}}
\email{yangzhang@nuaa.edu.cn}

\author{Zhaohui Yang}
\affiliation{%
  \institution{Zhejiang University}
  \city{Zhejiang}
  \country{China}}
\email{yang\_zhaohui@zju.edu.cn}

\author{Dusit Niyato}
\affiliation{%
  \institution{Nanyang Technological University}
  \city{Singapore}
  \country{Singapore}}
\email{dniyato@ntu.edu.sg}

\renewcommand{\shortauthors}{Jinbo Wen et al.}

\begin{abstract}
Web 3.0 represents the next generation of the Internet, which is widely recognized as a decentralized ecosystem that focuses on value expression and data ownership. By leveraging blockchain and artificial intelligence technologies, Web 3.0 offers unprecedented opportunities for users to create, own, and monetize their content, thereby enabling User-Generated Content (UGC) to an entirely new level. However, some self-interested users may exploit the limitations of content curation mechanisms and generate low-quality content with less effort, obtaining platform rewards under information asymmetry. Such behavior can undermine Web 3.0 performance. To this end, we propose \textit{LMM-Incentive}, a novel Large Multimodal Model (LMM)-based incentive mechanism for UGC in Web 3.0. Specifically, we propose an LMM-based contract-theoretic model to motivate users to generate high-quality UGC, thereby mitigating the adverse selection problem from information asymmetry. To alleviate potential moral hazards after contract selection, we leverage LMM agents to evaluate UGC quality, which is the primary component of the contract, utilizing prompt engineering techniques to improve the evaluation performance of LMM agents. Recognizing that traditional contract design methods cannot effectively adapt to the dynamic environment of Web 3.0, we develop an improved Mixture of Experts (MoE)-based Proximal Policy Optimization (PPO) algorithm for optimal contract design. Simulation results demonstrate the superiority of the proposed MoE-based PPO algorithm over representative benchmarks in the context of contract design. Finally, we deploy the designed contract within an Ethereum smart contract framework, further validating the effectiveness of the proposed scheme.
\end{abstract}




\maketitle

\section{Introduction}
Web 3.0 is a vision for the next generation of the Internet, grounded in decentralization, blockchain technology, and token-based economics, and designed to empower users with greater control over their data and digital identity~\cite{11082313, 10078899, wu2025web3}. Unlike the current Web 2.0, which is dominated by large corporations such as Meta and Google, Web 3.0 has the potential to revolutionize the contemporary Internet in two key dimensions~\cite{shen2024artificial}. From a data management perspective, Web 3.0 fosters a secure and trustworthy ecosystem for the management and seamless exchange of data, tokens, and digital assets~\cite{11082313}, which are linked to the blockchain and stored in distributed data storage systems~\cite{XiaoxuWeb3.0}. From a data ownership perspective, Web 3.0 diminishes corporate dominance. Without third-party authorities, users retain ownership of their unique digital identities built on blockchain technology to access personal data~\cite{XiaoxuWeb3.0}, thereby encouraging participation in a more user-centric and transparent online environment~\cite{10273397}.

In Web 3.0, the integration of Artificial Intelligence (AI) and blockchain technologies has initiated a paradigm shift in User-Generated Content (UGC). Rather than being confined to traditional content creation tools, users can now leverage AI-driven tools (e.g., generative AI models) to generate content~\cite{shen2024artificial}. One of the most significant advantages of Web 3.0 for UGC lies in content ownership and control. In contrast to traditional web platforms, where users often relinquish ownership rights and control once their creations are uploaded~\cite{11082313}, the decentralized, blockchain-empowered infrastructure of Web 3.0 allows users to establish verifiable ownership and retain full control over their generated content~\cite{XiaoxuWeb3.0}. Moreover, the immutable and decentralized nature of blockchain ensures that content remains tamper-proof and resistant to censorship~\cite{JianaWeb, JinboWeb3}. In addition, the tokenization capability of Web 3.0 has revolutionized the monetization of UGC~\cite{11082313}. Users can transform their digital creations into unique non-fungible tokens, which serve as verifiable representations of ownership over specific pieces of content~\cite{shen2024artificial}.

Generating content is generally costly and resource-intensive for some users, especially when relying on traditional generation methods~\cite{shen2024artificial}. To encourage users to participate in content generation, tokenized incentive platforms (e.g., Decentralized Autonomous Organizations (DAOs)) offer rewards to users for their contributions through transparent smart contracts~\cite{li2024attention, XiaoxuWeb3.0}. However, the private effort of users is unobservable to the platform~\cite{11091502}, leading to information asymmetry between the platform and users. Building on this, some self-interested users may exploit the limitations of content curation mechanisms, which rely on community voting and physical evaluation, to generate low-quality content with less effort~\cite{GANGWAR202181}. For example, users may coordinate voting rings or take advantage of the time lag between content submission and community evaluation to promote low-effort contributions, thus obtaining undeserved rewards and ultimately undermining the overall quality of the Web 3.0 ecosystem~\cite{shen2024artificial}.

To address the issue of information asymmetry, in this paper, we propose \textit{LMM-Incentive}, a novel Large Multimodal Model (LMM)-based incentive mechanism for UGC in Web 3.0. Specifically, we propose an LMM-based contract-theoretic model to incentivize users to generate high-quality UGC. Users can select suitable contract items according to their effort levels (i.e., the amount of resources). Nevertheless, some self-interested users may reduce their actual effort after contract acceptance while still claiming rewards, thereby creating moral hazards. To alleviate potential moral hazards after contract selection, some researchers quantify participant effort by incorporating physical equations or measurable metrics into the utility function of participants~\cite{7995145, 10570933, YinqiuReputation}. However, this approach fails to precisely capture the actual effort exerted by participants. To address this limitation, we employ LMM agents to directly evaluate UGC quality, which serves as the primary component of the contract. Evidently, traditional contract design approaches, which rely on complete prior information of the environment, cannot effectively adapt to the dynamic nature of Web 3.0~\cite{Wensus, KangTCCN}. Therefore, we develop an improved Mixture of Experts (MoE)-based Proximal Policy Optimization (PPO) algorithm for optimal contract design. Here, PPO, as one of the most representative Deep Reinforcement Learning (DRL) algorithms, has been widely applied across various domains owing to its efficiency and flexibility~\cite{RuichenMoE-PPO, JinboPPO}. Our contributions are summarized as follows:
\begin{itemize}
    \item \textbf{Novel LMM-Incentive for High-quality UGC:} We propose a novel LMM-based contract-theoretic model to motivate users to generate high-quality UGC in Web 3.0. Specifically, we first define user reputation values as continuous user types, and then design a contract with multiple items. Each user can select a contract item that best matches its effort level, thereby mitigating the adverse selection problem caused by information asymmetry. To address the moral hazard issue after contract selection, we develop and employ LMM agents to directly evaluate UGC quality, enhancing their evaluation capabilities by utilizing few-shot and Chain-of-Thought (CoT) prompting techniques, thereby effectively discouraging users from making low-effort contributions in content generation.
    \item \textbf{Improved MoE-PPO for Optimal Contract Design:} We develop an improved MoE-based PPO algorithm to obtain optimal contracts. Specifically, we integrate an MoE architecture into the PPO policy. The MoE architecture consists of a gating network and multiple expert networks. The gating network first calculates the weight probabilities of experts and selects the most suitable ones. The selected experts are then trained to design contracts based on feature data of environment states, while the gating network aggregates their outputs according to the calculated weights, thereby generating the final optimal contract.
    \item \textbf{Ethereum Platform Experiments with Optimal Contracts:} We first conduct extensive experiments to show the superior performance of the proposed MoE-based PPO algorithm compared with five representative benchmarks, i.e., PPO~\cite{JinboPPO}, transformer-based PPO~\cite{Transformer-PPO}, tiny PPO~\cite{Tiny_PPO}, Soft Actor-Critic (SAC)~\cite{haarnoja2018soft}, and the Generative Diffusion Model (GDM)~\cite{GDM}. Our algorithm achieves the highest train, test, and final rewards among all benchmarks, demonstrating the effectiveness and superiority of the proposed MoE-based PPO algorithm in optimal contract design. Furthermore, we deploy the designed contract within an Ethereum smart contract framework called Remix IDE\footnote{Remix IDE is available at {\url{https://remix.ethereum.org}}}, further validating the practical applicability of the proposed scheme.
\end{itemize}

The rest of the paper is organized as follows: Section \ref{Related_work} reviews the related work. Section \ref{Problem_Formulation} proposes the LMM-based contract-theoretic model. In Section \ref{MoE-PPO-Optimal-Contract-Design}, we develop the improved MoE-based PPO algorithm for optimal contract design. Section \ref{Simulation-Results} conducts a comprehensive evaluation of the proposed algorithm and scheme. Finally, Section \ref{Conclusion} concludes this paper.

\section{Related Work}\label{Related_work}
In this section, we provide a review of the related work across three domains, i.e., Web 3.0 and UGC, contract theory for provision, and DRL for contract design.

\subsection{Web 3.0 and UGC}
Owing to its potential to revolutionize Web 2.0, Web 3.0 has recently garnered significant attention from both academia and industry~\cite{11082313, shen2024artificial, 10273397}. In~\cite{11082313}, the authors presented a comprehensive review of Web 3.0, outlining its current landscape and recent advancements, including its architecture, enabling technologies, and ecosystem. In~\cite{10273397}, the authors proposed a Web 3.0 architecture designed to enable a zero-touch and zero-trust environment. In addition, more researchers have explored novel technologies to advance the development of Web 3.0~\cite{XiaoxuWeb3.0, JianaWeb, JinboWeb3, 10078899}. For instance, the authors in~\cite{JianaWeb} proposed a reliable block propagation optimization framework using graph attention networks for blockchain-enabled Web 3.0. In~\cite{XiaoxuWeb3.0}, the authors examined how quantum information technologies can enhance blockchain, thereby opening new avenues for the advancement of Web 3.0. UGC has become a driving force in the evolution of the Internet, empowering users to actively engage with and contribute to online platforms~\cite{shen2024artificial, 11091502}. In~\cite{11091502}, the authors proposed a three-player Bayesian persuasion game for social network platforms to suppress user-generated misinformation. In Web 3.0, low-quality UGC, such as malicious or deceptive content, can severely undermine the stability and trustworthiness of the ecosystem~\cite{shen2024artificial}. While the authors in~\cite{shen2024artificial} emphasized that AI technologies can play a significant role in detecting harmful UGC, our objective is to curb such behavior at its source. Therefore, it is essential to design a reliable incentive mechanism to encourage users to generate high-quality content in Web 3.0.

\begin{table*}[t]
\centering
\caption{Comparison Between the Current Works and Our Work in Contract Design.}\label{Related_work_comparison}
\footnotesize
\begin{tabular}
{p{2cm} >{\centering\arraybackslash}p{2.3cm} >{\centering\arraybackslash}p{2.3cm} 
                >{\centering\arraybackslash}p{2.3cm} >{\centering\arraybackslash}p{2.3cm}}
\toprule
\cellcolor{LightGray}\textbf{Literature} & \cellcolor{LightBlue}\textbf{Adverse Selection} & \cellcolor{LightBlue}\textbf{Moral Hazard} & \cellcolor{LightYellow}\textbf{Large Models} & \cellcolor{LightYellow}\textbf{DRL} \\ 
\midrule
\cite{6517781, 8239591, KangTCCN, 10841438}   & $\checkmark$ & $\times$ & $\times$ & $\times$ \\
\cite{10570933, 7995145}   & $\times$ & $\checkmark$ & $\times$ & $\times$ \\
\cite{YinqiuReputation}   & $\times$ & $\checkmark$ & $\times$ & $\checkmark$ \\
\cite{Wensus, wen2025diffusion, WenIoTJ, JinboPPO, ZhongContract}   & $\checkmark$ & $\times$ & $\times$ & $\checkmark$ \\
\cite{zhan2025learning}   & $\times$ & $\checkmark$ & $\checkmark$ & $\times$ \\
\cite{10925877}   & $\checkmark$ & $\times$ & $\checkmark$ & $\times$ \\
\midrule
\rowcolor{blue!5}
\textbf{Our work} & $\checkmark$ & $\checkmark$ & $\checkmark$ & $\checkmark$ \\
\bottomrule
\end{tabular}
\end{table*}

\subsection{Contract Theory for Provision}
Contract theory is a branch of economics that examines how individuals and organizations design contracts under conditions of information asymmetry~\cite{guruganesh2021contracts}. Due to its effectiveness, contract theory has been widely applied in wireless networks to ensure high-quality and reliable provision, including content~\cite{10570933}, data~\cite{KangTCCN, 10841438, Wensus, JinboPPO, 7995145}, services~\cite{YinqiuReputation, WenIoTJ, 10925877, zhan2025learning, wen2025diffusion}, and resources~\cite{6517781, 8239591, ZhongContract}. For instance, the authors in~\cite{KangTCCN} proposed a contract-theoretic model under Prospect Theory (PT) to motivate high-quality sensing data sharing in healthcare metaverses. In~\cite{YinqiuReputation}, the authors applied contract theory to optimize payment scheme design of clients, thereby ensuring the provision of high-quality AI-generated content services. In addition, in~\cite{ZhongContract}, the authors proposed a multi-dimensional contract-theoretic model under PT to incentivize roadside units to contribute bandwidth and computing resources, thereby enabling the seamless migration of embodied agent AI twins. However, the aforementioned works fail to simultaneously address the adverse selection and moral hazard problems arising from information asymmetry. Most importantly, \textit{how to design a contract-theoretic model for high-quality content provision remains an open problem, particularly in the context of Web 3.0, where research on this aspect is still lacking.} For clarity, Table \ref{Related_work_comparison} summarizes the comparison of the current works and our work in contract design under asymmetric information. 

\subsection{DRL for Contract Design}
DRL is an AI technique that combines deep learning and RL, enabling agents to learn decision-making policies by interacting with environments~\cite{haarnoja2018soft, schulman2017proximal}. Since traditional contract design methods, such as heuristic optimization algorithms~\cite{6517781, 8239591, KangTCCN}, often struggle in practical environments where complete information is unavailable, recent studies have explored the use of DRL algorithms to design optimal contracts~\cite{JinboPPO}. In~\cite{JinboPPO}, the authors adopted PPO algorithms to determine optimal contracts. Moreover, GDMs possess strong generative capabilities that make them well-suited for dynamic network optimization~\cite{GDM}. Therefore, some researchers have developed diffusion-based DRL algorithms for optimal contract design~\cite{Wensus, wen2025diffusion, WenIoTJ, ZhongContract}. For instance, the authors in~\cite{WenIoTJ} employed a diffusion-based SAC algorithm to generate optimal contracts under PT. Building on this, the authors in~\cite{Wensus} proposed a sustainable diffusion-based SAC algorithm that incorporates dynamic structured pruning techniques to reduce the number of parameters in diffusion-based actor networks, thereby identifying optimal feasible contracts. However, GDMs still face challenges such as high computational costs, difficulties in debugging, and inconsistent performance in low-dimensional environments. Motivated by the successful integration of MoE and PPO~\cite{RuichenMoE-PPO}, in this paper, we propose an improved MoE-based PPO algorithm for optimal contract design.

\section{Problem Formulation}\label{Problem_Formulation}
In this section, we first introduce the lifecycle of UGC in Web 3.0 and highlight one of its potential risks. We then propose an LMM-based contract-theoretic model to address this risk.

\begin{figure*}[t]
    \centering
    \includegraphics[width=0.95\textwidth]{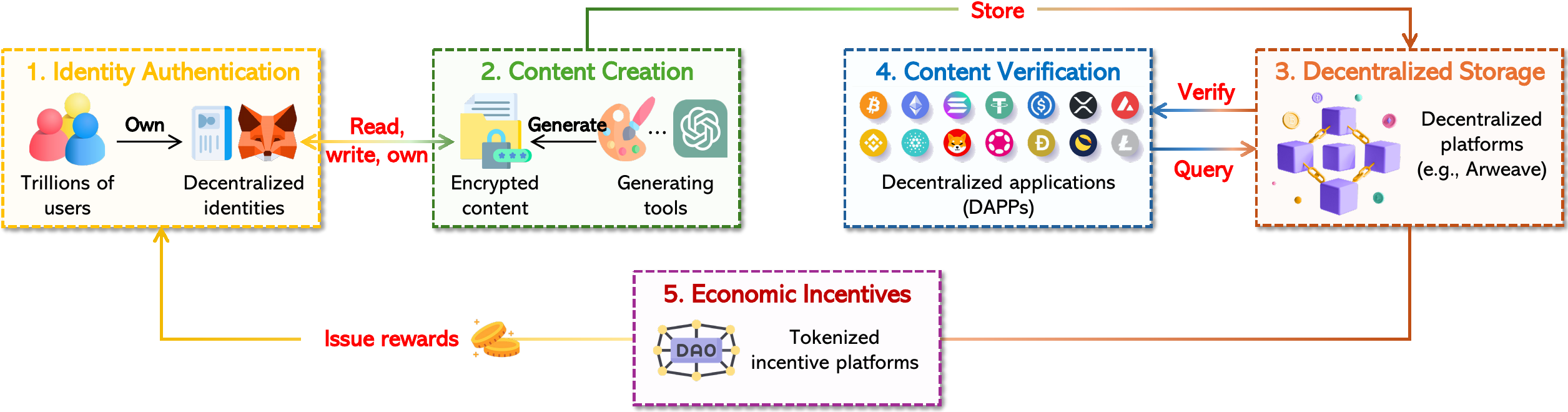}
    \caption{The lifecycle of UGC in Web 3.0, mainly consisting of five steps: identity authentication, content creation, decentralized storage, content verification, and economic incentives.}
    \Description{A flow diagram showing the UGC process.}
    \label{UGC}
\end{figure*}

\subsection{UGC Lifecycle}
In Web 3.0, users have control over their own generated content, without any single entity monopolizing the network infrastructure or user information~\cite{JinboWeb3}. As shown in Fig. \ref{UGC}, the lifecycle of UGC in Web 3.0 is presented as follows:
\begin{itemize}
    \item \textbf{Step 1. Identity Authentication:} In Web 3.0, users first generate cryptographic key pairs and establish decentralized identities~\cite{11082313}. The key pairs consist of public and private keys, where the public key functions as an on-chain identifier and the private key enables signing and authorization~\cite{11082313}. Additionally, users can leverage privacy-preserving cryptographic methods, such as zero-knowledge proofs, to establish verifiable claims without the need for centralized intermediaries.
    \item \textbf{Step 2. Content Creation:} Users can liberally generate content with the assistance of AI tools or entirely through autonomous AI agents under predefined parameters~\cite{shen2024artificial}. Once creation is completed, the content is hashed, and the hash is digitally signed using the private key of users, binding the content to their decentralized identities and creating an immutable proof of authorship and integrity.
    \item \textbf{Step 3. Decentralized Storage:} The UGC is uploaded to decentralized storage platforms~\cite{11082313}, while metadata, including the hash and authorship details, is registered on the blockchain for performance and traceability. Moreover, users disseminate their generated content through peer-to-peer networks for easy retrieval~\cite{JianaWeb}.
    \item \textbf{Step 4. Content Verification:} Users in the peer-to-peer network can verify content authenticity by checking the signature against the public key of creators and matching the stored hash. During verification, smart contracts can automatically manage authorization and execute rules, such as access control, following the pre-set terms.
    \item \textbf{Step 5. Economic Incentives:} In Web 3.0 ecosystems, tokenized incentive platforms (e.g., DAOs) can directly reward users for their contributions to generating content~\cite{li2024attention}, either by distributing tokens from community treasuries or through smart contracts that encode predefined incentive rules~\cite{11082313}.
\end{itemize}

The primary purpose of economic incentive design is to align user participation with value creation. However, tokenized incentive platforms may lack awareness of the current effort level of users, resulting in a situation of information asymmetry. Some self-interested users would deliberately generate low-quality content to serve their interests, disrupting the global interests of the Web 3.0 ecosystem~\cite{10273397}. Therefore, it is necessary to design a reliable incentive mechanism to curb this phenomenon, thereby ensuring the sustainable operation of Web 3.0~\cite{10273397}.

\subsection{LMM-based Contract-Theoretic Model}
Contract theory is an effective economic tool to address the issue of information asymmetry~\cite{KangTCCN}, which focuses on designing contracts that motivate agents to act in ways that benefit both sides, even when they have different goals~\cite{WenIoTJ}. \textit{Adverse selection} and \textit{Moral hazard} are two fundamental information asymmetries in contract theory~\cite{guruganesh2021contracts}. Specifically, adverse selection refers to a situation in which agents, possessing private information unknown to the principal, tend to select biased contracts to obtain more benefits at the expense of the principal. To mitigate this issue, the principal must design incentive-compatible contracts~\cite{guruganesh2021contracts}. Moral hazard arises after contract selection, when the principal may find it difficult to directly supervise agent actions~\cite{guruganesh2021contracts}. Although the principal can employ metrics to indirectly evaluate performance, agents may still resort to opportunistic behaviors that deviate from the interest of the principal.

Most of the current research focuses on addressing one of the information asymmetries, either adverse selection~\cite{WenIoTJ, ZhongContract, Wensus} or moral hazard~\cite{7995145, 10570933, YinqiuReputation}. To simultaneously address these two forms of information asymmetries, researchers employ physical equations to quantify agent effort and incorporate them into the utility function of agents~\cite{10078899}. However, this way still cannot objectively and exactly quantify the real effort of agents~\cite{10925877}. Since UGC in Web 3.0 is inherently multimodal~\cite{11082313}, and LMMs demonstrate strong capabilities in analyzing multimodal data~\cite{HongyangTPAMI}, we develop and employ LMM agents to quantify the generation effort of users and propose an LMM-based contract-theoretic model to promote high-quality UGC in Web 3.0. In the following, we formulate the utilities of the reward platform and its users, respectively.

\subsubsection{User utility}
The user utility depends on the received rewards from the platform and the cost of generating content. In particular, the cost of generating content depends on the quality of content~\cite{WenIoTJ}, that is, the higher the quality, the greater the generation cost. Moreover, the quality of UGC is generally associated with user reputation~\cite{YinqiuReputation}. Once users multiply issue low-quality content, their on-chain reputation values will be decreased. Thus, we denote the quality of UGC as $\mathcal{Q}(\phi)$ and utilize it to approximate the cost~\cite{10925877}, where $\phi \in [\underline{\phi}, \overline{\phi}]$ represents user reputation. The utility of users can be expressed as
\begin{equation}
    \mathcal{U}_U = f\phi R(\phi) - \kappa \mathcal{Q}(\phi),
\end{equation}
where $R(\phi)$ represents the received rewards associated with the user reputation, $f$ is a pre-defined weight parameter, and $\kappa$ is the unit cost of generating content~\cite{WenIoTJ}.

\subsubsection{Reward platform utility} The utility of the reward platform depends on the quality of UGC and the rewards given to users. Specifically, high-quality content will attract more users to consume tokens, thereby enhancing the value and liquidity of the platform token~\cite{11082313}. Therefore, the utility of the reward platform toward each user is expressed as 
\begin{equation}
    \mathcal{U}_P = \eta\ln ((\mathcal{Q}(\phi) - I) + 1) - R(\phi),
\end{equation}
where $\eta$ is the coefficient utilized to fine-tune the utility of the reward platform~\cite{10925877}, and $I$ represents the threshold of UGC quality.

\begin{figure*}[t]
    \centering
    \includegraphics[width=0.82\textwidth]{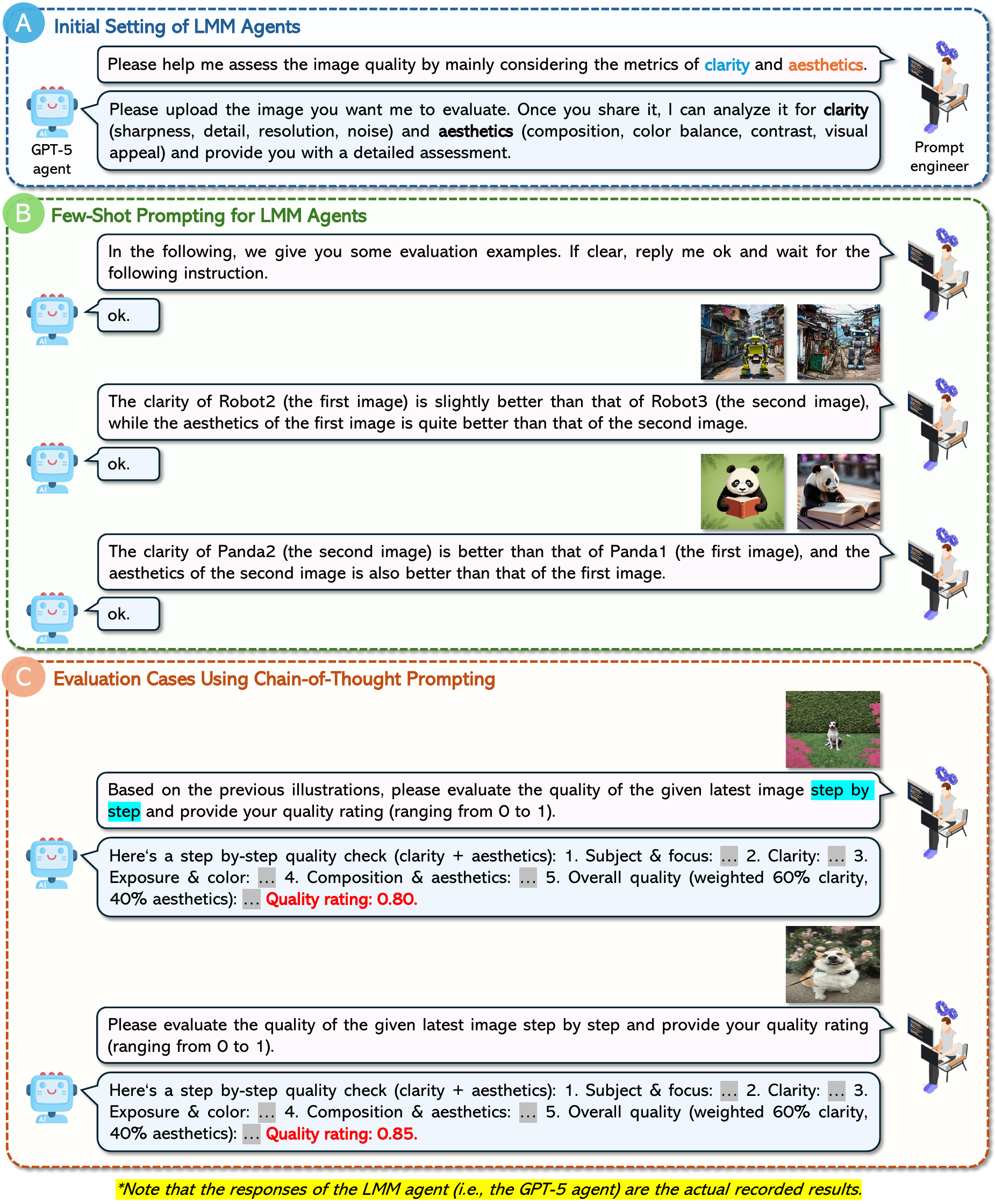}
    \caption{A procedure of UGC evaluation through an LMM agent. \textbf{Part A} illustrates the initial setting of the LMM agent. \textbf{Part B} introduces few-shot prompting to enhance its performance of UGC evaluation. \textbf{Part C} presents evaluation cases conducted by the LMM agent, where CoT prompting is employed to guide it to generate accurate results. Note that in practical implementation, we append the prompt \textit{``Please directly output the quality rating''} to accelerate the UGC evaluation.}
    \Description{A procedure of UGC evaluation through LMM agents.}
    \label{LMMContract}
\end{figure*}

\subsubsection{LMM-based contract formulation} 
In practice, the platform can observe only the past on-chain reputation values of users~\cite{11082313}, but cannot precisely determine their current effort levels and reputation status, indicating that there exists an information asymmetry between the platform and users. Therefore, we define the user reputation $\phi$ as the user type, which is a continuous variable. In the LMM-based contract-theoretic model, the platform, acting as the principal, is responsible for designing a contract consisting of a series of items, denoted by $\Omega = \{(\mathcal{Q}(\phi), R(\phi))\}$. Each user in the platform selects the appropriate contract item according to its type before content generation, which can mitigate the problem of adverse selection~\cite{WenIoTJ}.

In addition to employing multiple indicators to indirectly calculate $\mathcal{Q}(\phi)$, the platform leverages trained LMM agents to directly evaluate the quality of UGC and obtain a precise value of $\mathcal{Q}(\phi)$. As shown in Fig. \ref{LMMContract}, we present a procedure of UGC evaluation through an LMM agent (i.e., a GPT-5 agent). Specifically, we utilize the LMM agent to evaluate the quality of images generated by Stable Diffusion\footnote{Stable Diffusion is available at {\url{https://huggingface.co/spaces/stabilityai/stable-diffusion}}}, which mainly considers two factors: clarity and aesthetics~\cite{HongyangTPAMI}. Similarly, we can employ LMM agents to evaluate the quality of other multimodal content, such as text, by incorporating the corresponding factors. To enhance the performance of the LMM agent, we first apply few-shot prompting by providing a small set of evaluation examples~\cite{brown2020language}, which enhances its adaptability and reasoning capability. After this initialization, we utilize CoT prompting to guide the LMM agent to conduct step-by-step evaluation~\cite{wei2022chain}. Through prompt engineering, the LMM agent is thereby enabled to generate more accurate evaluation results $\mathcal{Q}(\phi)$. Guided by LMM agent feedback, the platform can monitor user effort in content generation, thereby alleviating the problem of moral hazard. To guarantee that each user automatically chooses the contract item designed for its specific type, the feasible contract should satisfy the following \textit{Individual Rationality (IR)} and \textit{Incentive Compatibility (IC)} constraints~\cite{WenIoTJ, 10841438, KangTCCN}.

\begin{definition}\label{IR_constraint}
    \textnormal{IR Constraints:} Users obtain a non-negative utility when they select the contract item corresponding to their types, i.e.,
    \begin{equation}\label{IR}
        \mathcal{U}_U = f\phi R(\phi) - \kappa \mathcal{Q}(\phi) \geq 0.
    \end{equation}
\end{definition}

\begin{definition}\label{IC_constraint}
    \textnormal{IC Constraints:} Users achieve the maximum utility by selecting the contract item designed for their types rather than any other item, i.e.,
    \begin{equation}\label{IC}
        f\phi R(\phi) - \kappa \mathcal{Q}(\phi) \geq  f\phi R(\hat{\phi}) - \kappa \mathcal{Q}(\hat{\phi}),\:\: \phi, \hat{\phi} \in [\underline{\phi}, \overline{\phi}].
    \end{equation}
\end{definition}

According to the above IR and IC constraints, we formulate the problem of maximizing the expected utility of the reward platform as follows:
\begin{equation}\label{problem1}
    \begin{aligned}
        \textbf{Problem 1:} \: & \max_{(\mathcal{Q}(\phi), R(\phi))} \int_{\underline{\phi}}^{\overline{\phi}} [\eta\ln ((\mathcal{Q}(\phi) - I) + 1) - R(\phi)] \mathcal{F}(\phi) \, \mathrm{d}\phi\\
        &\:\:\quad\:\:\text{s.t.}\:\: f\phi R(\phi) - \kappa \mathcal{Q}(\phi) \geq 0,\\
        &\quad\:\:\quad\:\:\:\: f\phi R(\phi) - \kappa \mathcal{Q}(\phi) \geq  f\phi R(\hat{\phi}) - \kappa \mathcal{Q}(\hat{\phi}),\\
        &\quad\:\:\quad\:\:\:\: \mathcal{Q}(\phi) \geq I,\: \phi, \hat{\phi} \in [\underline{\phi}, \overline{\phi}],
    \end{aligned}
\end{equation}
where $\mathcal{F}(\phi)$ represents the probability density function of user reputation $\phi$. According to~\cite{deshprabhu2024identification}, node reputation in distributed networks follows the beta distribution. Thus, we model user reputation $\phi$ as a beta-distributed random variable, and $\mathcal{F}(\phi)$ is given by
\begin{equation}
    \mathcal{F}(\phi) = \frac{\Gamma(\alpha + \beta)}{\Gamma(\alpha)\Gamma(\beta)}\phi^{\alpha-1}(1-\phi)^{\beta-1},\: \alpha,\beta > 0,
\end{equation}
where $\Gamma(\cdot)$ is the gamma function, and $\alpha$ and $\beta$ are the shape parameters of the distribution.

Obviously, it is challenging to directly solve Problem 1 under the existing IR (\ref{IR}) and IC (\ref{IC}) constraints. A common approach is to reduce the number of IR and IC constraints to decrease the solution complexity~\cite{WenIoTJ, KangTCCN}. Thus, we first reduce the number of IR constraints.

According to the IC constraints, we have
\begin{equation}
    f\phi R(\phi) - \kappa \mathcal{Q}(\phi) \geq  f\phi R(\underline{\phi}) - \kappa \mathcal{Q}(\underline{\phi}) \geq  f\underline{\phi} R(\underline{\phi}) - \kappa \mathcal{Q}(\underline{\phi}).
\end{equation}
If the IR constraint of $\underline{\phi}$ is satisfied, the IR constraints for all the other values of $\phi$ will automatically hold. Thus, the IR constraints can be simplified as
\begin{equation}
    f\underline{\phi} R(\underline{\phi}) - \kappa \mathcal{Q}(\underline{\phi}) \geq 0.
\end{equation}

To reduce the number of IC constraints, $\mathcal{Q}(\phi)$ is required to satisfy the monotonicity condition~\cite{10841438, 6517781}. However, users with high reputation values may also unintentionally generate low-quality content. Therefore, $\mathcal{Q}(\phi)$ does not strictly satisfy the monotonicity condition. Based on the above analysis, Problem 1 can be simplified as
\begin{equation}\label{problem2}
    \begin{aligned}
        \textbf{Problem 2:} \: & \max_{(\mathcal{Q}(\phi), R(\phi))} \int_{\underline{\phi}}^{\overline{\phi}} [\eta\ln ((\mathcal{Q}(\phi) - I) + 1) - R(\phi)] \mathcal{F}(\phi) \, \mathrm{d}\phi\\
        &\:\:\quad\:\:\text{s.t.}\:\: f\underline{\phi} R(\underline{\phi}) - \kappa \mathcal{Q}(\underline{\phi}) \geq 0,\\
        &\quad\:\:\quad\:\:\:\: f\phi R(\phi) - \kappa \mathcal{Q}(\phi) \geq  f\phi R(\hat{\phi}) - \kappa \mathcal{Q}(\hat{\phi}),\\
        &\quad\:\:\quad\:\:\:\: \mathcal{Q}(\phi) \geq I,\: \phi, \hat{\phi} \in [\underline{\phi}, \overline{\phi}].
    \end{aligned}
\end{equation}

The solution to Problem 2 mainly proceeds in two stages. The first is to calculate $\mathcal{Q}(\phi)$ using LMM agents, and the second is to derive the optimal reward $R^{\star}(\phi)$ based on the calculated $\mathcal{Q}(\phi)$, ensuring that the utility of the platform can be maximized under the IR and IC constraints. Since contracts can be more easily broadcast in a limited number of forms~\cite{6517781}, we quantize the continuous distribution of $\phi$ with a discrete one consisting of a finite set of points. Specifically, we first quantize the type set $[\underline{\phi}, \overline{\phi}]$ using a quantization factor $K$, thereby reducing it to a discrete set of $K$ types, denoted as $\Phi = \{\phi_1,\phi_2,\ldots,\phi_K\}$. The quantization process is uniform with equidistant values~\cite{6517781}, i.e., $\phi_k = \underline{\phi} + \frac{k-1}{K}(\overline{\phi} - \underline{\phi}),\: k\in [1,\ldots, K]$, sorted as $\phi_1 < \phi_2 < \cdots < \phi_K$. Additionally, the probability of users belonging to type $\phi_k$ is formulated by applying the forward difference method, which can be expressed as~\cite{6517781}
\begin{equation}
    \delta_k = \mathcal{P}(\phi_k \leq \phi \leq \phi_{k+1}) = 
    \left\{
    \begin{array}{l}
        \widetilde{\mathcal{F}}(\phi_{k+1}) - \widetilde{\mathcal{F}}(\phi_k), \: 1 \leq k < K, \\
        \widetilde{\mathcal{F}}(\overline{\phi}) - \widetilde{\mathcal{F}}(\phi_K), \:\quad k = K, \\
    \end{array}
    \right.
\end{equation}
where we have $\sum_{k = 1}^K \delta_k =1$, and $\widetilde{\mathcal{F}}(\cdot)$ represents the cumulative distribution function of $\phi$, following the beta distribution. Therefore, Problem 2 can be reformulated as
\begin{equation}\label{problem3}
    \begin{aligned}
        \textbf{Problem 3:} \: & \max_{(\mathcal{Q}(\phi_k), R(\phi_k))}\Bigg[\mathbb{E}(\mathcal{U}_P) = \sum_{k = 1}^K \delta_k (\eta\ln ((\mathcal{Q}(\phi_k) - I) + 1) - R(\phi_k))\Bigg]\\
        &\:\:\quad\:\:\:\text{s.t.}\:\: f\phi_1 R(\phi_1) - \kappa \mathcal{Q}(\phi_1) \geq 0,\\
        &\quad\:\:\:\quad\:\:\:\: f\phi_k R(\phi_k) - \kappa \mathcal{Q}(\phi_k) \geq  f\phi_k R(\phi_j) - \kappa \mathcal{Q}(\phi_j),\\
        &\quad\:\:\:\quad\:\:\:\: \phi_1 < \phi_2 < \cdots < \phi_K,\\
        &\quad\:\:\:\quad\:\:\:\: \mathcal{Q}(\phi_k) \geq I,\: k, j \in [1, K].
    \end{aligned}
\end{equation}

Although the number of IR constraints is reduced, it is still difficult to use traditional optimization approaches to solve Problem 3 under the existing IC constraints. Moreover, in practice, environment parameters such as $\eta$ and $\kappa$ may change over time, necessitating the redesign of traditional approaches~\cite{Wensus}. Hence, traditional optimization approaches may not provide an effective solution to Problem 3. DRL algorithms have been widely applied to network optimization problems, as they can learn policies in dynamic environments with incomplete information~\cite{WenIoTJ, Wensus}, and achieve near-optimal solutions in scenarios where traditional methods often struggle.

\begin{figure*}[t]
    \centering
    \includegraphics[width=0.95\textwidth]{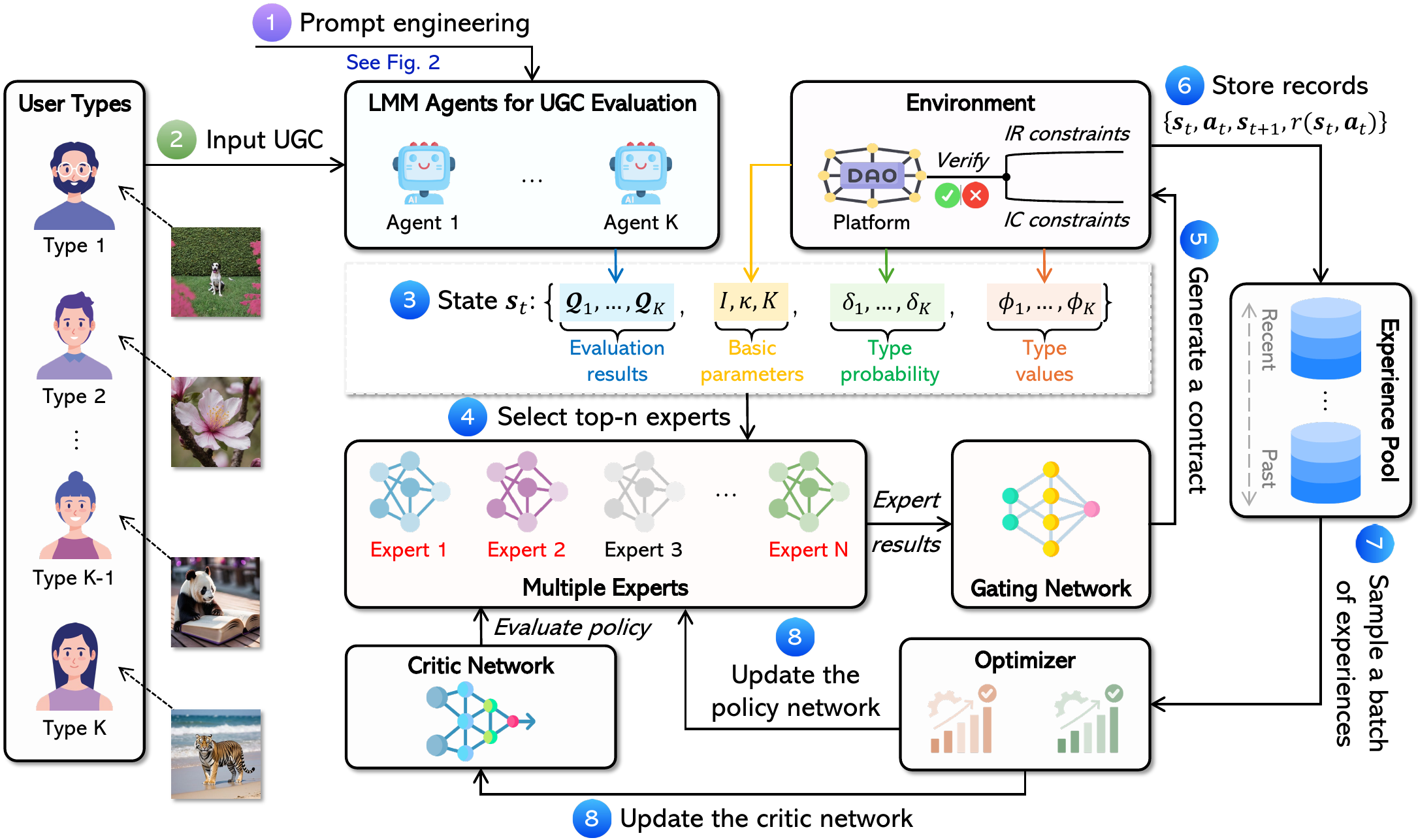}
    \caption{The diagram of the MoE-based PPO algorithm for optimal contract design. LMM agents are first enhanced through prompt engineering (\textbf{Step 1}), after which they evaluate the quality of the input UGC (\textbf{Step 2}). In \textbf{Steps 3 to 8}, the training process of the MoE-based PPO algorithm is presented.}
    \Description{MoE-based PPO for optimal contract design.}
    \label{MoE-PPO}
\end{figure*}

\section{Improved MoE-based PPO for Optimal Contract Design}\label{MoE-PPO-Optimal-Contract-Design}
In this section, we introduce the improved MoE-based PPO algorithm for optimal contract design. We first model Problem 3 as a Markov Decision Process (MDP). We then present the architecture of the proposed MoE-based PPO algorithm, as shown in Fig. \ref{MoE-PPO}. It is worth noting that the MoE-based PPO algorithm is implemented on the reward platform to optimize contract design.

\subsection{MDP Formulation}
In the MDP framework, we formulate Problem 3 through the components $\langle \mathcal{S}, \mathcal{A}, \mathcal{R}, \gamma \rangle$, where $\mathcal{S}$ is the state space consisting of state vectors $\mathbf{s}$, $\mathcal{A}$ is the action space consisting of action vectors $\mathbf{a}$, $\mathcal{R}$ is the reward function, and $\gamma \in [0,1]$ is the discount factor governing the trade-off between present and future returns. We detail these components as follows:

\subsubsection{State vector} The derivation of optimal contracts begins with LMM agents quantifying $\mathcal{Q}(\phi)$, followed by a DRL agent that learns the optimal reward $R^{\star}(\phi)$ based on the estimated $\mathcal{Q}(\phi)$. Thus, we incorporate $\mathcal{Q}(\phi)$ into the state $\mathbf{s}$, which is expressed as
\begin{equation}
   \mathbf{s} \triangleq \{\mathcal{Q}(\phi_1),\ldots,\mathcal{Q}(\phi_K), I, \kappa, K, \delta_1,\ldots,\delta_K, \phi_1,\ldots,\phi_K\},
\end{equation}
where the total dimensionality of the state $\mathbf{s}$ is $(3K + 3)$.

\subsubsection{Action vector} 
The objective of the DRL agent is to determine the optimal reward $R^{\star}(\phi)$ conditioned on the state $\mathbf{s}$. Hence, the action of the DRL agent $\mathbf{a}$ is given by
\begin{equation}
   \mathbf{a} \triangleq \{R(\phi_1),\ldots,R(\phi_K)\},
\end{equation}
where the total dimensionality of the action $\mathbf{a}$ is $K$.

\subsubsection{Reward function}
The reward function plays a significant role in guiding the DRL agent to maximize the expected utility of the reward platform while satisfying the constraints in (\ref{problem3}). Thus, the reward function $\mathcal{R}$ for the DRL agent is defined as
\begin{equation}
    \mathcal{R} = r(\mathbf{s}, \mathbf{a}) = 
    \left\{
    \begin{array}{l}
        \mathbb{E}(\mathcal{U}_P), \: \text{if constraints in (\ref{problem3}) are satisfied}, \\
        0, \:\quad\quad\: \text{otherwise}. \\
    \end{array}
    \right.
\end{equation}

\subsection{MoE-based PPO Architecture}
PPO is one of the most representative DRL algorithms, which has been widely applied across various domains~\cite{JinboPPO}. Notably, PPO lays the foundation for large language model alignment owing to its stable training dynamics and strong adaptability in complex environments~\cite{wen2024reinforcing}. However, when handling complex tasks, PPO requires relatively high computational demands, and limited resources can adversely affect its convergence speed. MoE is a machine learning architecture composed of a gating network and multiple sub-networks, referred to as \textit{experts}~\cite{jiayihe}. Each expert is trained to specialize in a subset of the input data, while the gating network selectively activates only a subset of experts for each input, thereby enabling high model capacity with efficient computation~\cite{chen2022towards}. In particular, MoE layers can be broadly categorized into two types: dense MoE and sparse MoE. A dense MoE layer activates all experts simultaneously, whereas a sparse MoE layer activates only a selected subset of experts.

Recently, MoE has been integrated into PPO to enhance the capacity of the policy~\cite{RuichenMoE-PPO, wen2024defending}. For instance, a dense MoE-based PPO algorithm was proposed to solve the sum-rate maximization problem in heterogeneous satellite networks~\cite{RuichenMoE-PPO}. Motivated by this, we propose an improved MoE-based PPO algorithm, which adopts an actor-critic framework, to design optimal contracts. Specifically, the actor network, parameterized by $\boldsymbol{\theta}_a$, is augmented with an ensemble of experts to facilitate decision-making, while the critic network, parameterized by $\boldsymbol{\theta}_c$, assesses the state-value function $V_{\boldsymbol{\theta}_c}(\mathbf{s})$ to guide policy improvement.

\subsubsection{MoE-based actor network}
We consider an MoE-based actor network consisting of $M$ expert networks, denoted as $\mathcal{E}_{\boldsymbol{\psi}_1}, \mathcal{E}_{\boldsymbol{\psi}_2}, \ldots, \mathcal{E}_{\boldsymbol{\psi}_M}$, where each expert is a perception with unique parameters $\boldsymbol{\psi}_i$. To select the most relevant experts based on the current state feature $\mathbf{x}$, the central gating network first generates unnormalized expert weights $g_{\boldsymbol{\varphi}}(\mathbf{x}) \in \mathbb{R}^M$, as given by
\begin{equation}
    g_{\boldsymbol{\varphi}}(\mathbf{x}) = \mathbf{W}_g \mathbf{x} + \mathbf{b}_g,
\end{equation}
where $\mathbf{W}_g$ is a learnable weight matrix, and $\mathbf{b}_g$ is a bias vector. Then, these expert weights $g_{\boldsymbol{\varphi}}(\mathbf{x})$ are normalized through the softmax function to produce the expert selection probabilities, expressed as
\begin{equation}
    p_{\boldsymbol{\varphi}, i} = \frac{\exp(g_{\boldsymbol{\varphi}, i}(\mathbf{x}))}{\sum_{j = 1}^M \exp(g_{\boldsymbol{\varphi}, j}(\mathbf{x}))},\: i = 1, 2, \ldots, M.
\end{equation}

In the sparse setting, instead of activating all experts, the gating network selects only the top-$m$ experts with the highest probabilities, where $m \leq M$. Specifically, the weights corresponding to the top-$m$ experts are retained and re-normalized, while those of the remaining experts are set to $0$, i.e.,
\begin{equation}\label{probability_normalized}
    p_{\boldsymbol{\varphi}, i}^{\prime} = \left\{
    \begin{array}{l}
        \frac{p_{\boldsymbol{\varphi}, i}}{\sum_{j\in \text{top-}m}p_{\boldsymbol{\varphi}, j}}, \: i \in \text{top-}m, \\
        0, \:\quad\quad\: \text{otherwise}. \\
    \end{array}
    \right.
\end{equation}

Based on (\ref{probability_normalized}), the mean of the MoE-based policy is given by
\begin{equation}
    \mu_{\boldsymbol{\theta}_a}(\mathbf{s}) = \sum_{i = 1}^M p_{\boldsymbol{\varphi}, i}^{\prime} \mathcal{E}_{\boldsymbol{\psi}_i} (\mathbf{x}).
\end{equation}
To enable more stable training, we define the standard deviation as $\sigma_{\boldsymbol{\theta}_a}(\mathbf{s}) = \exp(\boldsymbol{\omega}_{\sigma})$, where $\boldsymbol{\omega}_{\sigma} \in \mathbb{R}^{|\mathbf{a}|}$ are independent learnable parameters. Therefore, the policy network of the proposed MoE-based PPO framework is given by
\begin{equation}
    \pi_{\boldsymbol{\theta}_a}(\mathbf{a}|\mathbf{s}) = \mathcal{N}(\mathbf{a} \mid \mu_{\boldsymbol{\theta}_a}(\mathbf{s}), \Sigma_{\boldsymbol{\theta}_a}(\mathbf{s})),
\end{equation}
where $\Sigma_{\boldsymbol{\theta}_a}(\mathbf{s})$ is a diagonal covariance matrix, expressed as $\mathrm{diag}\big(\sigma^2_{\boldsymbol{\theta}_a}(\mathbf{s})\big)$.

\subsubsection{Objective function of MoE-based PPO}
According to~\cite{RuichenMoE-PPO}, the objective function of the proposed MoE-based PPO algorithm preserves the core principles of PPO, which is formulated as~\cite{schulman2017proximal}
\begin{equation}\label{PPO_objective}
    \begin{aligned}
        & \max_{\boldsymbol{\theta}_a} \Bigg[\mathcal{L}_A(\boldsymbol{\theta}_a) = \mathbb{E}_{t}\bigg[\frac{\pi_{\boldsymbol{\theta}_a}(\mathbf{a}_t|\mathbf{s}_t)}{\pi_{\boldsymbol{\theta}^{\mathrm{old}}_a}(\mathbf{a}_t|\mathbf{s}_t)} A^{\pi_{\boldsymbol{\theta}^{\mathrm{old}}_a}}(\mathbf{s}_t, \mathbf{a}_t)\bigg] \Bigg]\\
        &\:\:\:\text{s.t.}\: \mathbb{E}_{t}\big[D_{\mathrm{KL}}(\pi_{\boldsymbol{\theta}_a}(\mathbf{a}_t|\mathbf{s}_t) \| \pi_{\boldsymbol{\theta}^{\mathrm{old}}_a}(\mathbf{a}_t|\mathbf{s}_t))\big] \leq \varepsilon,
    \end{aligned}
\end{equation}
where $D_{\mathrm{KL}}(\cdot)$ is the Kullback-Leibler (KL) divergence function, and $\varepsilon$ is a positive hyperparameter~\cite{RuichenMoE-PPO}. In practice, the expectation $\mathbb{E}_{t}(\cdot)$ is approximated by an empirical average $\hat{\mathbb{E}}_{t}(\cdot)$ over a batch of sampled trajectories. $A^{\pi_{\boldsymbol{\theta}^{\mathrm{old}}_a}}(\mathbf{s}_t, \mathbf{a}_t)$ is the variance-reduced advantage function, which can be estimated by using the Generalized Advantage Estimation (GAE) based on the state-value function $V_{\boldsymbol{\theta}_c}(\mathbf{s})$, which can be formulated as~\cite{schulman2017proximal} 
\begin{equation}
   A^{\pi_{\boldsymbol{\theta}^{\mathrm{old}}_a}}(\mathbf{s}_t, \mathbf{a}_t) = \sum_{\ell = 0}^{T-t-1}(\gamma \lambda)^{\ell} \Delta_{t+\ell},\: \Delta_t = r(\mathbf{s}_t, \mathbf{a}_t) + \gamma V_{\boldsymbol{\theta}^{\mathrm{old}}_c}(\mathbf{s}_{t+1}) - V_{\boldsymbol{\theta}^{\mathrm{old}}_c}(\mathbf{s}_t),
\end{equation}
where $T$ is the maximum step in an episode, and $\lambda \in [0, 1]$ is a balance factor that trades off bias and variance in advantage estimation. In particular, when $\lambda = 1$, GAE reduces to the Monte Carlo estimation of the advantage function, expressed as~\cite{JinboPPO, schulman2017proximal}
\begin{equation}
  A^{\pi_{\boldsymbol{\theta}^{\mathrm{old}}_a}}(\mathbf{s}_t, \mathbf{a}_t) = \gamma^{T-t} V_{\boldsymbol{\theta}^{\mathrm{old}}_c}(\mathbf{s}_T) - V_{\boldsymbol{\theta}^{\mathrm{old}}_c}(\mathbf{s}_t) + \sum_{\ell = t}^{T-1} \gamma^{\ell-t} r(\mathbf{s}_\ell, \mathbf{a}_\ell).
\end{equation}

To facilitate computation, the objective function in (\ref{PPO_objective}) is directly constrained to ensure that the update from $\boldsymbol{\theta}^{\mathrm{old}}_a$ to $\boldsymbol{\theta}_a$ is not excessively large, rather than considering the KL divergence constraint~\cite{RuichenMoE-PPO}. Thus, the objective function $\mathcal{L}_A(\boldsymbol{\theta}_a)$ can be approximated by~\cite{schulman2017proximal}
\begin{equation}
    \mathcal{L}_A^{\mathrm{CLIP}}(\boldsymbol{\theta}_a) = \mathbb{E}_{t}\bigg[\min\bigg(\frac{\pi_{\boldsymbol{\theta}_a}(\mathbf{a}_t|\mathbf{s}_t)}{\pi_{\boldsymbol{\theta}^{\mathrm{old}}_a}(\mathbf{a}_t|\mathbf{s}_t)} A^{\pi_{\boldsymbol{\theta}^{\mathrm{old}}_a}}(\mathbf{s}_t, \mathbf{a}_t),\mathrm{clip}\bigg(\frac{\pi_{\boldsymbol{\theta}_a}(\mathbf{a}_t|\mathbf{s}_t)}{\pi_{\boldsymbol{\theta}^{\mathrm{old}}_a}(\mathbf{a}_t|\mathbf{s}_t)}, 1 - \epsilon, 1 + \epsilon \bigg) A^{\pi_{\boldsymbol{\theta}^{\mathrm{old}}_a}}(\mathbf{s}_t, \mathbf{a}_t)\bigg) \bigg], 
\end{equation}
where $\mathrm{clip}(\cdot)$ is the clip function that restricts the probability ratio to the range $[1-\epsilon, 1+\epsilon]$, with $\epsilon$ being a predefined hyperparameter.

\subsubsection{Update mechanisms of policy and critic}
In our MoE-based PPO implementation, the parameters of both the actor and critic networks are optimized by minimizing their respective loss functions. Specifically, the loss function of the critic network is given by~\cite{schulman2017proximal}
\begin{equation}
    \mathcal{L}_C(\boldsymbol{\theta}_c) = \mathbb{E}_t [\|V_{\boldsymbol{\theta}_c}(\mathbf{s}_t) - V_{\mathrm{tar}}(\mathbf{s}_t) \|^2],
\end{equation}
where $V_{\mathrm{tar}}(\mathbf{s}_t)$ is the target return estimated from the trajectory, as given by~\cite{RuichenMoE-PPO}
\begin{equation}
    V_{\mathrm{tar}}(\mathbf{s}_t) = r(\mathbf{s}_t, \mathbf{a}_t) + \cdots + \gamma^{\ell + 1} V_{\boldsymbol{\theta}^{\mathrm{old}}_c}(\mathbf{s}_{t + \ell + 1}).
\end{equation}
Thus, the parameters of the critic network are updated through mini-batch Stochastic Gradient Descent (SGD) using experiences sampled from the environment~\cite{RuichenMoE-PPO}, expressed as
\begin{equation}\label{critic_update}
    \boldsymbol{\theta}_c \gets \boldsymbol{\theta}^{\mathrm{old}}_c - \tau_c \frac{1}{B}\sum_{t=1}^B \nabla_{\boldsymbol{\theta}_c} \mathcal{L}_C(\boldsymbol{\theta}_c),
\end{equation}
where $\tau_c \in (0,1]$ denotes the learning rate of the critic network, and $B$ is the mini-batch size.

To encourage balanced utilization of experts, we incorporate an MoE-specific auxiliary loss function into the total loss function of the actor network, which is formulated as
\begin{equation}
    \mathcal{L}_{\mathrm{MoE}}(\boldsymbol{\theta}_a) = - \mathbb{E}_t \bigg[\sum_{i=1}^M p_{\boldsymbol{\varphi}, i}^{\prime} \log p_{\boldsymbol{\varphi}, i}^{\prime} \bigg]. 
\end{equation}
Therefore, the total function of the actor network is expressed as
\begin{equation}
    \mathcal{L}_{A}^{\mathrm{Total}}(\boldsymbol{\theta}_a) = \mathcal{L}_A^{\mathrm{CLIP}}(\boldsymbol{\theta}_a) + \omega_{\mathrm{MoE}} \mathcal{L}_{\mathrm{MoE}}(\boldsymbol{\theta}_a) - \omega_{\mathrm{Ent}} \mathbb{E}_t\big[\mathcal{H}\big[\pi_{\boldsymbol{\theta}^{\mathrm{old}}_a}(\cdot|\mathbf{s}_t)\big]\big],
\end{equation}
where $\mathcal{H}\big[\pi_{\boldsymbol{\theta}^{\mathrm{old}}_a}(\cdot|\mathbf{s}_t)\big]$ denotes the entropy of the action probability distribution, encouraging exploration by discouraging overly deterministic policies~\cite{wen2024defending}, and $\omega_{\mathrm{MoE}}$ and $\omega_{\mathrm{Ent}}$ represent the weight coefficients of the MoE auxiliary loss and the entropy, respectively. Thus, the parameters of the actor network are updated by the following function:
\begin{equation}\label{actor_update}
        \boldsymbol{\theta}_a \gets \boldsymbol{\theta}^{\mathrm{old}}_a - \tau_a \frac{1}{B}\sum_{t=1}^B \nabla_{\boldsymbol{\theta}_a} \mathcal{L}_{A}^{\mathrm{Total}}(\boldsymbol{\theta}_a),
\end{equation}
where $\tau_a \in (0,1]$ denotes the learning rate of the actor network.

\begin{algorithm}[t]
\footnotesize
\label{diffusion_algorithm}
\DontPrintSemicolon
\SetAlgoLined

\textbf{Input:} The number of experts, auxiliary loss coefficient, learning rate, episodes, and environment steps.

Initialize contract environment and relay buffer $\mathcal{B}$.

\For{\rm{the episode} $e=1$ \rm{to} $E_{\mathrm{max}}$}
{

    Observe initial state $\mathbf{s}_0$.

\For{$t=1$ \rm{to} $T$}
{   
    \textcolor{blue}{\textit{\#\#\# Action generation with MoE \#\#\#}}

    Each expert generates its own action $\mathcal{E}_{\boldsymbol{\psi}_i} (\mathbf{x})$ based on the current state.

    Obtain the current reward $r(\mathbf{s}_t, \mathbf{a}_t)$. 
    
    \textcolor{blue}{\textit{\#\#\# Experience collections \#\#\#}}
    
    Observe the next state $\mathbf{s}_{t+1}$.

    Store record $(\mathbf{s}_t,\mathbf{a}_t,\mathbf{s}_{t+1}, r(\mathbf{s}_t, \mathbf{a}_t))$ into the relay buffer $\mathcal{B}$.

}

    \textcolor{blue}{\textit{\#\#\# Parameter updates \#\#\#}}
    
    Sample a random mini-batch of transitions with size $B$ from $\mathcal{B}$.

    Update the critic parameters $\boldsymbol{\theta}_c$ through SGD by using (\ref{critic_update}).
    
    Update the policy parameters $\boldsymbol{\theta}_a$ through SGD by using (\ref{actor_update}).

    Update the current state $\mathbf{s}_t \gets \mathbf{s}_{t + 1}$.
}
\textbf{Output:} The optimal reward $R^{\star}(\phi) = \{R^{\star}(\phi_1),\ldots,R^{\star}(\phi_K)\}$.

\caption{The proposed MoE-based PPO algorithm for optimal contract design}\label{Algorithm_MoE}
\end{algorithm}

The pseudocode of the proposed MoE-based PPO algorithm is presented in Algorithm \ref{Algorithm_MoE}. The computational complexity of Algorithm \ref{Algorithm_MoE} consists of two components: $\mathcal{C}_{\mathrm{MoE}}$ and $\mathcal{C}_{\mathrm{Critic}}$. On the one hand, the computational complexity of the MoE-based actor network mainly comes from the gating network and the experts, which is given by $\mathcal{O}(BM|\mathbf{x}|) + \mathcal{O}(Bm|\mathbf{x}||\mathbf{a}|)$. Here, $|\mathbf{x}|$ is the state feature dimensionality, and $|\mathbf{a}|$ is the action dimensionality. On the other hand, the computational complexity of the critic network is $\mathcal{O}(Bh^2)$~\cite{JinboPPO}, where $h$ is the hidden size of the critic network. Therefore, the total computational complexity of the proposed MoE-based PPO algorithm is $\mathcal{O}(B(M|\mathbf{x}| + m|\mathbf{x}||\mathbf{a}| + h^2))$.

\section{Simulation Results}\label{Simulation-Results}
In this section, we first provide the details of the experimental setup. We then evaluate the performance of the proposed MoE-based PPO in the context of optimal contract design. Finally, we implement the designed contract within an Ethereum smart contract framework.

\subsection{Experimental Setup}

\subsubsection{Experimental platform} We conduct experiments to evaluate the performance of the proposed MoE-based PPO algorithm for optimal contract design, implemented in PyTorch with CUDA 12.0 and executed on an NVIDIA GeForce RTX 3080 Laptop GPU. Furthermore, we deploy the designed contract within Remix IDE using Solidity 0.8.17. Remix IDE is an Ethereum-based smart contract development platform maintained by the Ethereum community, which provides an integrated environment for writing, compiling, debugging, and deploying smart contracts.

\subsubsection{Environment details} 
\begin{figure*}[t]
    \centering
    \includegraphics[width=0.95\textwidth]{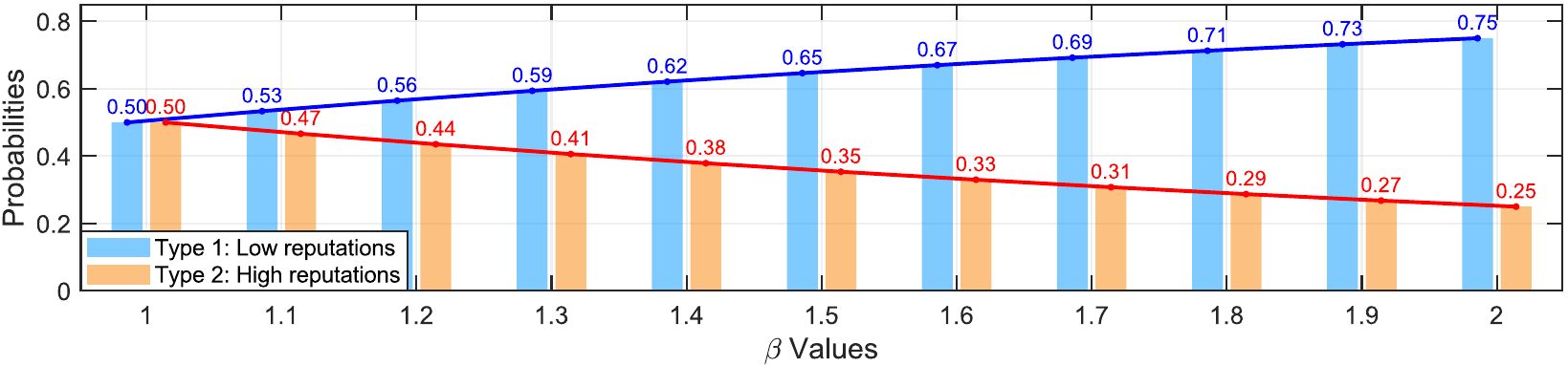}
    \caption{The probability functions for two user types with the shape parameter $\beta$ as the independent variable. We consider user reputation with two types: low and high~\cite{10925877}, and set the value of $\alpha$ to $1$. Note that, when $\beta$ is fixed, the probability function with the shape parameter $\alpha$ as the independent variable is symmetrical to the current probability function.}
    \Description{Type probability function.}
    \label{Type_probability}
\end{figure*}

For clarity, we consider user reputation with two types: low and high~\cite{10925877}, i.e., $K = 2$. Regarding the environment parameters, the threshold of UGC quality $I$ is set to $0$, the unit cost of generating content $\kappa$ is randomly sampled from the interval $[1, 3)$~\cite{wen2025diffusion}, the maximum user reputation value $\overline{\phi}$ and the minimum user reputation value $\underline{\phi}$ are randomly sampled within $[5,10)$ and $[10,15)$, respectively~\cite{YinqiuReputation}, and the shape parameters of the beta distribution $\alpha, \beta$ are both randomly sampled within $[1,2]$~\cite{deshprabhu2024identification}, where the probability functions for two user types are presented in Fig. \ref{Type_probability}. Moreover, to enhance training efficiency and reduce computational costs, we create an LMM simulator to generate evaluation results of UGC. During the test stage, the LMM simulator can be replaced with the real LMM agent (e.g., the GPT-5 agent) through the API.

\subsubsection{Algorithm design} 
For the actor network structure, both the gating network and the expert networks are implemented as linear layers in the context of our contract design, thereby reducing the computational complexity of the actor network. To effectively evaluate the policy, the critic network is designed with two fully-connected layers, each of size $256$, followed by $\mathrm{Tanh}$ activation functions~\cite{RuichenMoE-PPO}. The parameters of both the actor and critic networks are updated using the Adam optimizer~\cite{JinboPPO, RuichenMoE-PPO}, with a weight decay rate of $0.0001$ applied to regularize the model weights~\cite{wen2025diffusion}. For other hyperparameter settings, the batch size is set to $512$~\cite{JinboPPO}, the discount factor $\gamma$ to $0.95$~\cite{wen2024defending}, the value function coefficient to $0.5$~\cite{JinboPPO}, the clipping parameter $\epsilon$ to $0.2$~\cite{JinboPPO}, the maximum norm for gradient clipping to $0.5$~\cite{wen2024defending}, and the balance factor $\lambda$ to $0.95$~\cite{JinboPPO}.

\subsubsection{Benchmark comparisons}
We compare the proposed MoE-based PPO algorithm with three representative on-policy DRL algorithms, including PPO~\cite{schulman2017proximal}, transformer-based PPO~\cite{Transformer-PPO}, and tiny PPO~\cite{Tiny_PPO}, as well as two widely used off-policy DRL algorithms, including SAC~\cite{haarnoja2018soft} and the GDM~\cite{GDM}. We evaluate all algorithms in a discrete action space to ensure fair comparisons. Moreover, we compare the proposed scheme under asymmetric information with three schemes:
\begin{itemize}
    \item \textit{Contract scheme under complete information~\cite{8239591}:} The reward platform has access to user reputation information (i.e., user types) without considering the IC constraints (\ref{IC}).
    \item \textit{Random scheme under asymmetric information~\cite{WenIoTJ}:} The reward platform designs contracts randomly, without regard to user types.
    \item \textit{Average scheme under asymmetric information~\cite{ZhongContract}:} The reward platform designs identical contracts for all users, regardless of user types.
\end{itemize}

\subsection{Performance Evaluation of MoE-based PPO}
\begin{figure*}[t]
		\centering
		\begin{subfigure}[Scheme comparisons.]{\includegraphics[width=0.315\linewidth]{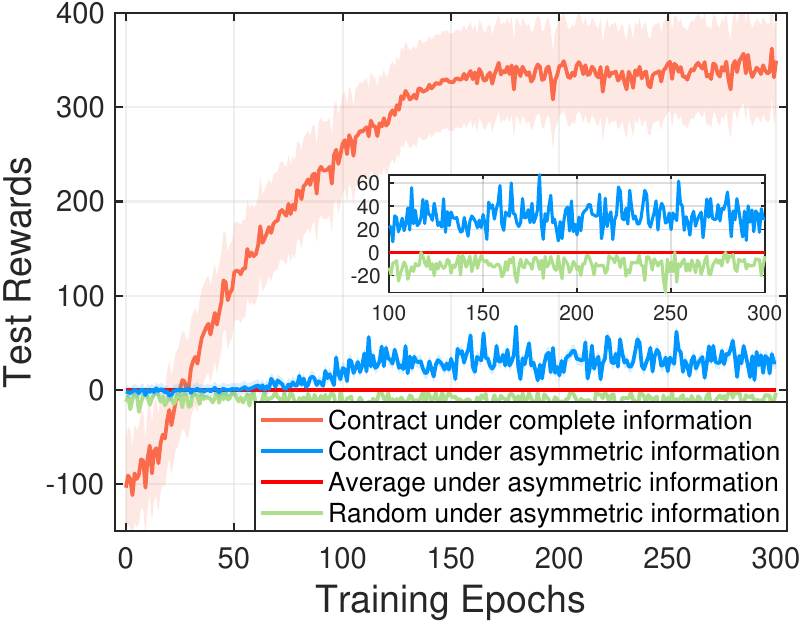}\label{Scheme_comparison}}
		\end{subfigure}
		\hfill
		\begin{subfigure}[On-policy comparisons.]{\includegraphics[width=0.3025\linewidth]{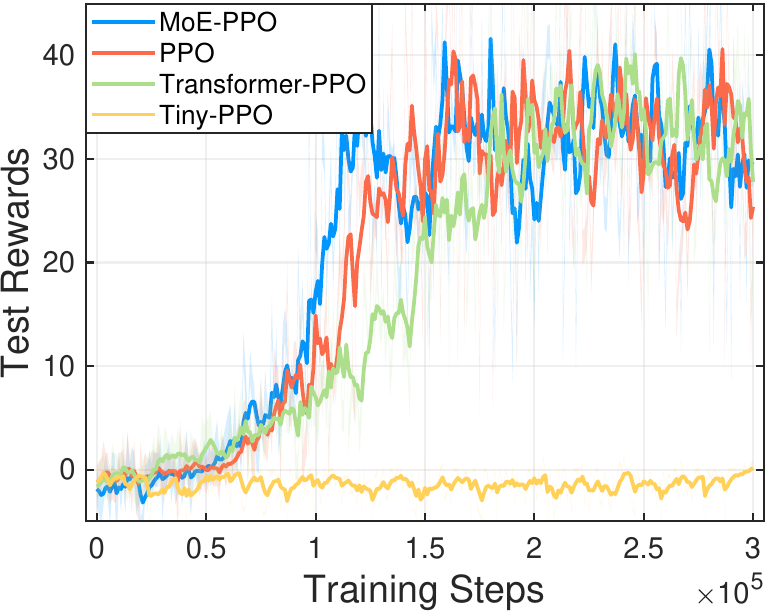}\label{PPO_comparison}}
		\end{subfigure}
		\hfill
		\begin{subfigure}[Off-policy comparisons.]{\includegraphics[width=0.3025\linewidth]{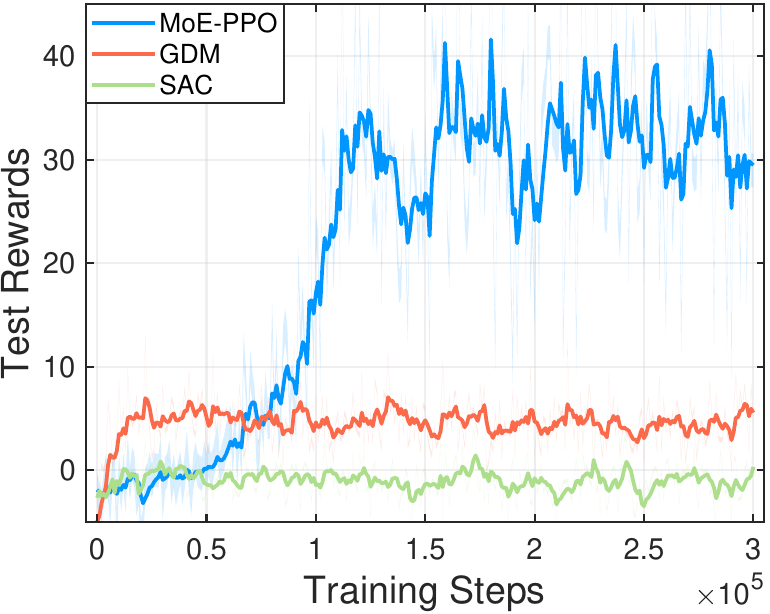}\label{AC_comparison}}
		\end{subfigure}
		\caption{Performance evaluation of the proposed MoE-based PPO algorithm in optimal contract design.}
        \Description{(a) Comparison of schemes. (b) Performance comparison of PPO. (c) Comparison of actor-critic methods.}
		\label{MoE-PPO_performance_evaluation}
\end{figure*}

In Fig. \ref{MoE-PPO_performance_evaluation}, we conduct the performance evaluation of the proposed MoE-based PPO algorithm in optimal contract design. From Fig. \ref{Scheme_comparison}, we observe that our contract scheme under asymmetric information outperforms the baseline schemes, namely the random and average schemes under asymmetric information. In addition, the contract scheme under complete information achieves significantly higher test rewards than the proposed contract scheme. The reason is that the reward platform has access to user type information and thus provides the most suitable contract item for each user~\cite{WenIoTJ}. However, in practice, it is difficult for the reward platform to precisely identify user type information, and a complete information environment is unrealistic. Therefore, the contract scheme under complete information can be seen as an ideal benchmark. Figures \ref{PPO_comparison} and \ref{AC_comparison} show performance comparisons of the proposed MoE-based PPO algorithm with three on-policy DRL algorithms and two off-policy DRL algorithms in the context of contract design. To present the comparison results more clearly, we summarize the performance achieved by the proposed MoE-based PPO algorithm and these DRL benchmarks in Table \ref{rl-comparison-with-benchmarks}, in terms of test rewards, train rewards, final rewards, total time, and convergence time. We see that the proposed MoE-based PPO algorithm achieves the highest test, train, and final rewards among all benchmarks. Moreover, it converges faster than PPO. These results are attributed to the collaborative ability of experts, demonstrating the superior characteristics of the proposed MoE-based PPO algorithm, including \textit{high performance} and \textit{fast convergence}.

\begin{table*}
		\centering
		\caption{Performance Comparisons of MoE-based PPO and Benchmarks.}
		\label{rl-comparison-with-benchmarks}
		\footnotesize
		\renewcommand\arraystretch{1.25}
		\begin{tabular}{c|c|ccccc}
			\toprule
            \rowcolor{gray!10}
			\multicolumn{2}{c|}{\textbf{Policy}} & \textbf{Tes. Reward} & \textbf{Tra. Reward} & \textbf{Fin. Reward} & \textbf{Tot. Time / h} & \textbf{Con. Time / h} \\  
			\hline\hline
			\multirow{2}{*}{Heuristic} & Random & $-10.32\pm5.81$ & $-$ & $-3.16\pm5.60$ & $0.0125$ & $\infty$ \\ 
			& Average & $0$ & $-$ & $0$ & $0.0136$ & $\infty$ \\ 
			\hline\hline
			\multirow{5}{*}{DRL} & PPO & $\underline{20.63}\pm16.16$ & $\underline{21.20}\pm14.57$ & $\underline{28.61}\pm10.28$ & $\textbf{0.2747}$ & $0.1606$ \\
			& Transformer-PPO & $18.56\pm15.07$ & $18.57\pm13.91$ & $18.91\pm8.66$ & $0.6325$ & $0.3478$ \\ 
			& Tiny-PPO & $-1.38\pm1.65$ & $-1.42\pm0.57$ & $0.53\pm1.69$ & $\underline{0.3936}$ & $0.3936+$ \\ 
			& GDM & $4.64\pm2.46$ & $4.65\pm0.99$ & $4.43\pm2.27$ & $4.2114$ & $\textbf{0.1353}$ \\ 
			& SAC & $-0.95\pm1.95$ & $0.69\pm0.61$ & $2.77\pm1.98$ & $1.5308$ & $0.1633$ \\ 
			\hline\hline
            \rowcolor{blue!5}
			\textbf{Ours} & \textbf{MoE-PPO} & $\textbf{21.75}\pm16.30$ & $\textbf{21.77}\pm13.87$ & $\textbf{28.93}\pm10.34$ & $0.4256$ & $\underline{0.1586}$ \\
			\bottomrule
		\end{tabular}
	\end{table*}

\begin{figure*}[t]
\centering
\subfigure[The impact of the total number of experts, with $\omega_{\mathrm{MoE}} = 0.01$, $m = 1$, and $\tau_a = 10^{-4}$.]
{
    \begin{minipage}[t]{0.38\linewidth}
	\centering
	\includegraphics[width=0.98\linewidth]{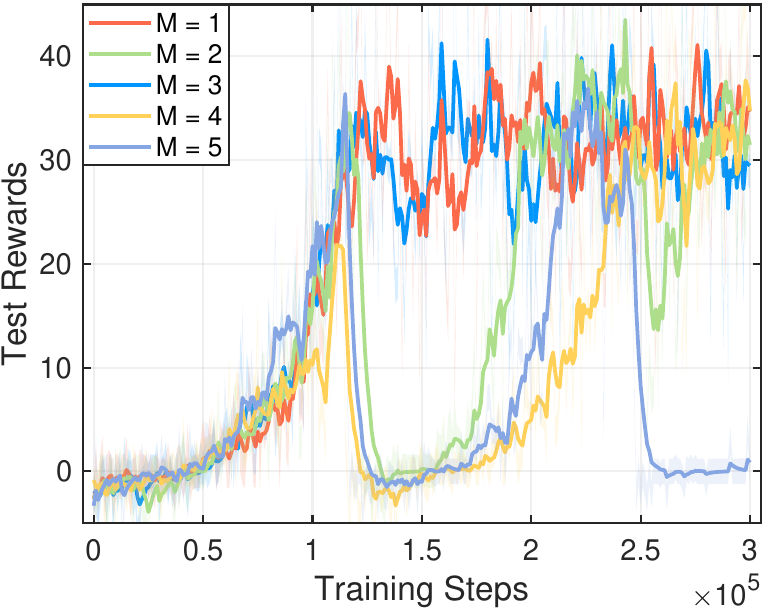}\label{MoE_num}
    \end{minipage}
}\hspace{0.5in}
\subfigure[The impact of the number of selected experts, with $M = 4$, $\omega_{\mathrm{MoE}} = 0.01$, and $\tau_a = 10^{-4}$.]
{
    \begin{minipage}[t]{0.38\linewidth}
	\centering
	\includegraphics[width=0.98\linewidth]{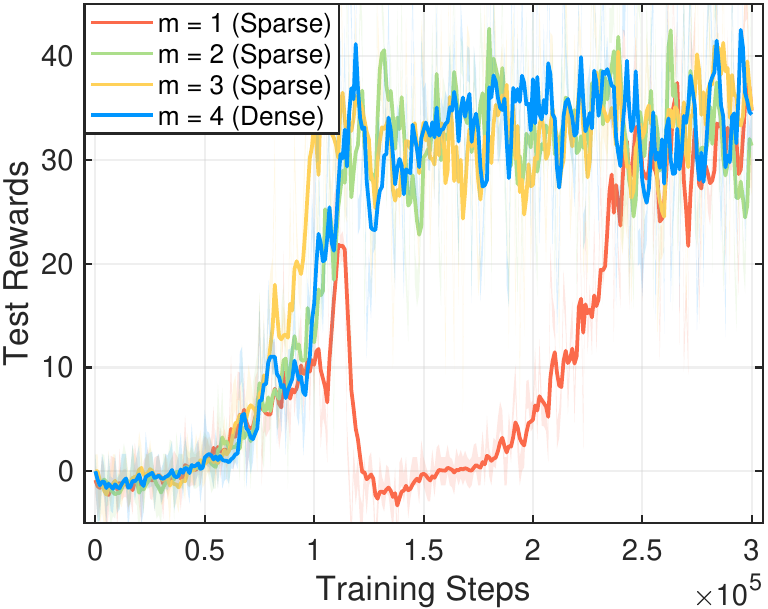}
	\label{MoE_m}
    \end{minipage}
}\hspace{0.5in}
\subfigure[The impact of auxiliary loss weights, with $M = 3$, $m = 1$, and $\tau_a = 10^{-4}$.]
{
    \begin{minipage}[t]{0.38\linewidth}
	\centering
	\includegraphics[width=0.98\linewidth]{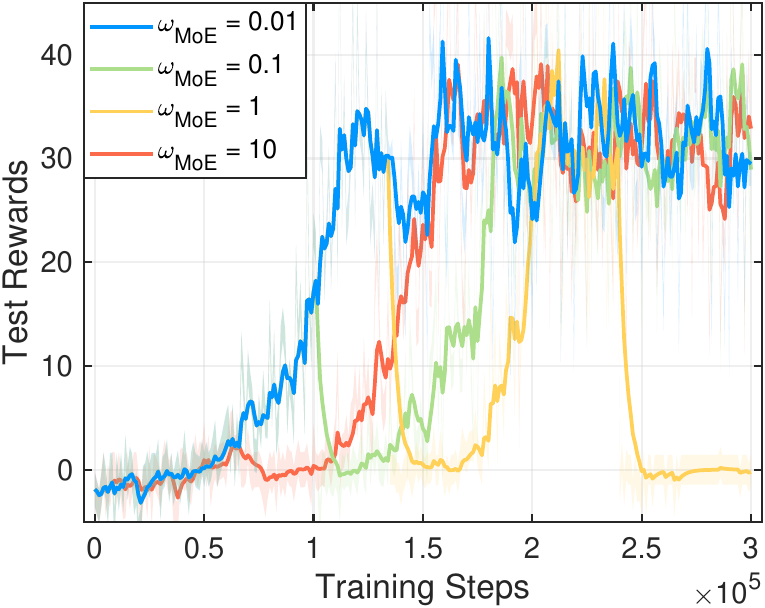}
	\label{MoE_aux}
    \end{minipage}
}\hspace{0.5in}
\subfigure[The impact of learning rates, with $M = 3$, $\omega_{\mathrm{MoE}} = 0.01$, and $m = 1$.]
{
    \begin{minipage}[t]{0.38\linewidth}
	\centering
	\includegraphics[width=0.98\linewidth]{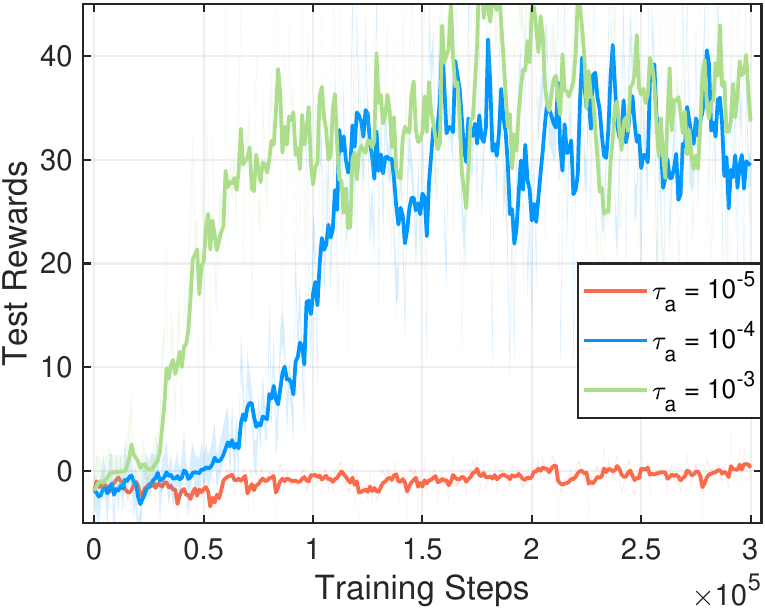}
	\label{MoE_lr}
    \end{minipage}
}
\caption{Performance evaluation of the proposed MoE-based PPO algorithm under hyperparameter settings}\label{MoE_PPO_hypertest}
\Description{Performance evaluation of the MoE-based PPO algorithm under different hyperparameter settings.}
\end{figure*}

In Fig. \ref{MoE_PPO_hypertest}, we present the performance of the proposed MoE-based PPO algorithm with different hyperparameter configurations. Specifically, Fig. \ref{MoE_num} illustrates the impact of the total number of experts $M$ on the performance of the MoE-based PPO algorithm, with $\omega_{\mathrm{MoE}} = 0.01$, $m = 1$, and $\tau_a = 10^{-4}$. We observe that, except for the case when $M = 5$, the MoE-based PPO algorithm converges for all other numbers of experts. This counterintuitive result can be attributed to the fact that an excessive number of experts increases the convergence difficulty of the gating network, which may lead to abnormal behavior or even the collapse of the policy. Figure \ref{MoE_m} shows the impact of the number of selected experts $m$ on the performance of the MoE-based PPO algorithm, with $M = 4$, $\omega_{\mathrm{MoE}} = 0.01$, and $\tau_a = 10^{-4}$. We observe that, when the MoE layer in our algorithm is sparse, increasing the number of selected experts can enhance the generalization ability of the policy, thereby improving the overall performance of the algorithm. Moreover, we find that the dense MoE layer in our algorithm reduces convergence speed because of the higher computational overhead from the gating mechanism. Figure \ref{MoE_aux} presents the impact of auxiliary loss weights $\omega_{\mathrm{MoE}}$ on the performance of the MoE-based PPO algorithm, with $M = 3$, $m = 1$, and $\tau_a = 10^{-4}$. We find that the MoE-based PPO algorithm achieves its best performance when $\omega_{\mathrm{MoE}} = 0.01$. The reason is that a small auxiliary loss weight can smoothly guide the gating network to aggregate expert outputs, while ensuring efficient policy learning. Figure \ref{MoE_lr} illustrates the impact of learning rates $\tau_a$ on the performance of the MoE-based PPO algorithm, with $M = 3$, $\omega_{\mathrm{MoE}} = 0.01$, and $m = 1$. We observe that our algorithm achieves the best performance with a learning rate of $\tau_a = 10^{-3}$. However, further increasing the learning rates may cause the policy to become unstable and eventually collapse, which has been tested.

\begin{figure*}[t]
    \centering
    \includegraphics[width=0.95\textwidth]{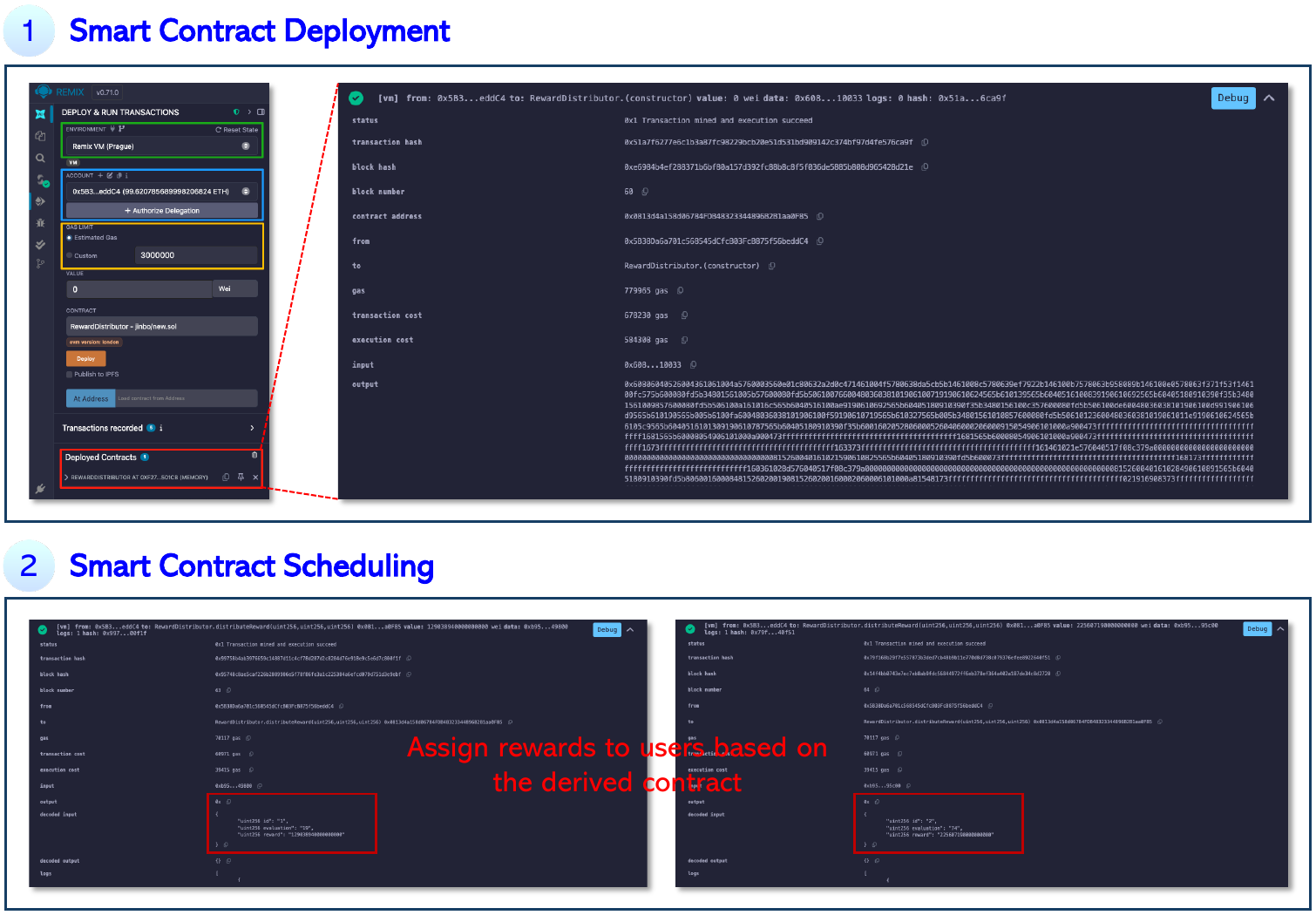}
    \caption{Deployment results of the near-optimal contract derived from the proposed MoE-based PPO algorithm within an Ethereum smart contract framework.}
    \Description{Smart contract implementation.}
    \label{smart_contract}
\end{figure*}

\subsection{Real Implementation of the Designed Contract}
As shown in Fig. \ref{smart_contract}, we deploy the near-optimal contract designed by the proposed MoE-based PPO algorithm within Remix IDE, which is an Ethereum smart contract framework. The smart contract is designed with five key variables: \textit{owner (address)}, \textit{id (uint256)}, \textit{recipient (address)}, \textit{evaluation (uint256)}, and \textit{reward (uint256)}. We describe the practical workflow of the smart contract as follows:
\begin{itemize}
    \item \textit{1. User identity mapping:} As shown in \textit{Part 1} of Fig. \ref{smart_contract}, we first initialize the smart contract by assigning a unique identifier to each user. The identifier is mapped to a verifiable wallet address on-chain, which is implemented through a public mapping structure: {\small \texttt{mapping(uint256 $\Rightarrow$ address) public idToAddress}}. This step ensures the traceability of users while decoupling real-world identity from blockchain identity, thereby preserving user privacy.
    \item \textit{2. Reward allocation:} As shown in \textit{Part 2} of Fig. \ref{smart_contract}, after deploying the contract designed by the proposed MoE-PPO algorithm into the smart contract, the corresponding reward value is determined and transferred directly from the owner to the wallet address of each user: {\small \texttt{recipient.transfer(reward)}}. The step guarantees that incentives are distributed in an automated and transparent manner without requiring third-party intervention.
    \item \textit{3. Contractual safeguards:} To prevent malicious exploitation of the designed contract, we enforce several security mechanisms, including a reentrancy lock that prevents reentrancy attacks, access control that restricts critical functions (e.g., reward distribution) to the owner only, and a reward allocation flag that introduces a state variable to record whether rewards have already been distributed to users, avoiding repeated transfers.
\end{itemize}

\section{Conclusion}\label{Conclusion}
In this paper, we have presented \textit{LMM-Incentive} for UGC in Web 3.0. Specifically, we have proposed an LMM-based contract-theoretic model to incentivize users to generate high-quality UGC under information asymmetry, thereby mitigating the adverse selection problem. To prevent users from exploiting the limitations of content curation mechanisms, we have developed and employed LMM agents to directly evaluate UGC quality, which is the primary item of the contract, utilizing prompt engineering techniques to enhance the evaluation performance of LMM agents, thereby alleviating potential moral hazards after contract selection. To obtain optimal contracts, we have developed an improved MoE-based PPO algorithm that integrates an MoE architecture into the actor network. Simulation results demonstrate the superior performance of the proposed MoE-based PPO algorithm compared with other benchmarks. Finally, we have deployed the designed contract within Remix IDE, further validating the effectiveness of the proposed scheme. For future work, we will employ real-world datasets to fine-tune LMMs for more accurate evaluation. In addition, we will extend the MoE-based PPO algorithm by integrating the transformer architecture, thereby better capturing the sequential dependencies among state features.

\bibliographystyle{ACM-Reference-Format}
\bibliography{sample-base}

\end{document}